\documentclass[11pt]{article}

\usepackage[final]{acl}

\usepackage{times}
\usepackage{latexsym}

\usepackage[T1]{fontenc}
\usepackage[utf8]{inputenc}
\usepackage{microtype}
\usepackage{inconsolata}

\usepackage{graphicx}
\usepackage[ruled,vlined,linesnumbered]{algorithm2e}
\usepackage{colortbl}
\usepackage[svgnames]{xcolor}
\usepackage{amsmath}
\usepackage{amssymb}
\usepackage{multirow}
\usepackage{booktabs}
\usepackage{pgfplots}
\usepackage[skins,breakable]{tcolorbox}
\usepackage{float}
\pgfplotsset{compat=1.18}

\definecolor{OpenSourceRow}{RGB}{245,245,245}
\definecolor{OurModelRow}{RGB}{238,238,238}
\definecolor{AvgHeader}{RGB}{238,238,238}
\definecolor{GroupHeader}{RGB}{243,246,250}
\definecolor{PromptBg}{RGB}{249,250,252}
\definecolor{PromptBorder}{RGB}{151,166,188}
\definecolor{PromptTitleBg}{RGB}{160,174,196}
\newcommand{\tablehead}[1]{\begin{tabular}[c]{@{}c@{}}\textbf{#1}\end{tabular}}
\newcommand{\tableheadtwo}[2]{\begin{tabular}[c]{@{}c@{}}\textbf{#1}\\\textbf{#2}\end{tabular}}
\newtcolorbox{promptbox}[1]{%
  enhanced,
  breakable,
  colback=PromptBg,
  colframe=PromptBorder,
  colbacktitle=PromptTitleBg,
  coltitle=black,
  fonttitle=\bfseries,
  title={#1},
  boxrule=0.6pt,
  arc=2pt,
  left=7pt,
  right=7pt,
  top=7pt,
  bottom=7pt,
  toptitle=3pt,
  bottomtitle=3pt,
  lefttitle=7pt,
  righttitle=7pt,
  before skip=6pt,
  after skip=6pt,
}
\newenvironment{compactitemize}
  {\begin{list}{$\bullet$}{%
    \setlength{\leftmargin}{1.45em}%
    \setlength{\labelwidth}{0.8em}%
    \setlength{\labelsep}{0.45em}%
    \setlength{\topsep}{0.05em}%
    \setlength{\partopsep}{0pt}%
    \setlength{\parsep}{0pt}%
    \setlength{\itemsep}{0.05em}%
  }}
  {\end{list}}

\graphicspath{{../figures/}{figures/}}

\title{Attend to Evidence: Evidence-Anchored Spatial Attention Supervision for Multimodal RLVR}

\author{
 \textbf{Ruina Hu\textsuperscript{1,2,†}},
 \textbf{Chen Wang\textsuperscript{2,4,†}},
 \textbf{Lai Wei\textsuperscript{2,5}},
 \textbf{Jionghao Bai\textsuperscript{2,6}},
 \textbf{Bin Yu\textsuperscript{1,2,3}},
\\
 \textbf{Weiran Huang\textsuperscript{5}},
 \textbf{Kai Wang\textsuperscript{1,*}},
 \textbf{Yue Wang\textsuperscript{2,7,*}}
\\
\\
 \textsuperscript{1}Harbin Institute of Technology \hspace{0.1cm}
 \textsuperscript{2}Zhongguancun Academy
\\
 \textsuperscript{3} Harbin Institute of Technology Qingdao Research Institute
\\
 \textsuperscript{4}Nankai University 
 \hspace{0.1cm}
 \textsuperscript{5}Shanghai Jiaotong University
 \hspace{0.1cm}
 \textsuperscript{6}Zhejiang University
 \hspace{0.1cm}
 \textsuperscript{7}XinZhuAI
\\
 \texttt{s-hrn25@bza.edu.cn, dr.wangkai@hit.edu.cn, yuewang@bza.edu.cn}
}

\begin{document}

\maketitle

\begingroup
\renewcommand{\thefootnote}{†}
\footnotetext{Equal contribution.}
\renewcommand{\thefootnote}{*}
\footnotetext{Equal corresponding authors.}
\endgroup

\begin{abstract}
Reinforcement learning with verifiable rewards (RLVR) improves vision-language models (VLMs) by optimizing outcome rewards derived from final answers. However, such outcome-only rewards do not tell the model which image regions justify an answer. 
For questions that require visual grounding, these rewards cannot distinguish responses supported by relevant visual evidence from those produced by language-prior shortcuts or lucky guesses.
We introduce \textbf{EASE} (Evidence-Anchored Spatial Attention), which augments multimodal RLVR with visual-evidence process supervision. EASE converts annotated evidence regions into a smoothed visual-token target and uses it to guide response-to-image attention during RL training, but only on high-reward trajectories. The annotations are used solely as privileged training labels, while inference requires only the original image and question.
Across Qwen2.5-VL-7B, Qwen3-VL-4B, and Qwen3-VL-8B, EASE raises average scores over DAPO by 2.5 to 3.1 points on perception, hallucination, visual math, and multimodal reasoning benchmarks. Diagnostics and ablations show that EASE better aligns visual attention with annotated evidence regions. 
Code are available at \url{https://github.com/Nrich-sunny/Attend-to-Evidence/}.
\end{abstract}

\begin{figure}[!t]
  \centering
  \includegraphics[width=\columnwidth]{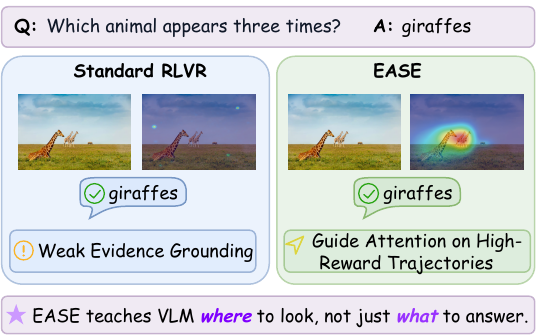}
  \caption{\textbf{From outcome rewards to evidence acquisition.} Standard RLVR may answer correctly while attending weakly to the key evidence region. EASE encourages stronger attention to the supporting region.}
  \label{fig:teaser}
\end{figure}

\section{Introduction}

Reinforcement learning with verifiable rewards (RLVR) has become a common way to improve language-model reasoning and is increasingly used in vision-language model (VLM) post-training~\citep{grpo,dapo,yang2025r1_onevision,liu2025visual_rft,tan2026reason_rft}. Outcome rewards are attractive in multimodal tasks because many answers can be checked automatically. Recent VLM systems show that these verifiable outcome signals can improve multimodal reasoning~\citep{wang2025thinklite_vl,vl-rethinker,meng2025mm_eureka,liu2025noisyrollout,wang2025papo,huang2025vppo,wang2026vgpo,cao2025ground_r1,ni2026point_rft}. However, outcome-only reward does not tell the model where the answer should come from in the image. A VLM may need to find a small object, compare two regions, or combine clues across separated regions before it can answer reliably. If this evidence-acquisition step is left unsupervised, RL can reward a correct answer without knowing whether the model actually used the relevant visual evidence rather than a shortcut.

This limitation creates a \textit{perception credit assignment} problem. An outcome reward tells the policy whether the response is correct, but not which image regions supported that response. A correct answer may reflect relevant visual evidence, but it may also come from language priors, dataset bias, a label-correlated shortcut, or guessing. An incorrect answer is also ambiguous because the model may have looked at the wrong region, or it may have found the right evidence and then reasoned incorrectly. This ambiguity matters most for fine-grained and multi-evidence questions that require object comparison, counting, localized attributes, spatial relations, or clues from separated regions. Without acquiring these regions, a model can produce plausible text while remaining weakly grounded and vulnerable to hallucination~\citep{li2023pope,guan2023hallusionbench,asadi2026mirage,shi2026vlmsseeingjustsaying,xu2026morethinking,fu2025hidden}.

Process supervision can reduce this ambiguity by making visual evidence acquisition part of the training signal. Evidence boxes and human attention annotations in visual question answering data indicate which image regions support an answer and can serve as training-only labels for where the model should look~\citep{hudson2019gqa,das2017vqa_hat}. Such supervision has been used to improve grounding in VLMs~\citep{selvaraju2019hint}. In RL, the same idea can be used to check more than the outcome. A high-reward response should also be supported by the visual evidence that justifies it. 
Recent multimodal RL and visual-thinking methods use stronger rollouts, perturbation objectives, visual-dependency signals, or textual traces augmented with grounding tags and spatial coordinates~\citep{liu2025noisyrollout,huang2025vppo,yuan2025visual,yang2025machine,zhu2026analyzing}. EASE instead uses evidence boxes only to supervise internal attention during training, leaving the VLM's inference format unchanged.

We propose \textbf{EASE} (\textbf{E}vidence-\textbf{A}nchored \textbf{S}patial Att\textbf{E}ntion), a framework that adds visual evidence-anchored process supervision to multimodal RLVR. Figure~\ref{fig:teaser} shows the intended contrast. Outcome-only RL may answer correctly while attending weakly to the image region that justifies the answer, whereas EASE encourages the model to keep the answer correct and place more attention on the supporting visual evidence.

EASE maps annotated evidence regions to a smoothed target distribution over visual tokens. During RL training, this target guides where generated response tokens attend within the image. The guidance is applied only to trajectories that receive high verifier rewards, since low-reward responses can have unreliable attention patterns. The evidence annotations are used only as privileged training labels, and inference uses the same image-question pathway as the underlying VLM without evidence metadata. The training data includes both examples with one supporting region and examples with multiple supporting regions, so the model learns to focus on a local cue when one region is sufficient and to cover several regions when the answer depends on multiple clues.

Empirically, EASE consistently improves outcome-only RL baselines across Qwen2.5-VL-7B, Qwen3-VL-4B, and Qwen3-VL-8B. When trained and evaluated under the same protocol as DAPO, EASE improves the average score by 2.9, 3.1, and 2.5 points on these backbones, respectively, with gains on perception-heavy reasoning, hallucination robustness, visual math, and logic benchmarks. EASE is also competitive with recent 7B-scale multimodal RL methods on shared benchmarks. Diagnostics and ablations link these gains to stronger evidence-aligned visual attention, reward-aware gating, the smoothed target, and mixed single-/multi-evidence training setup.

Our contributions are summarized as follows.
\begin{itemize}
    \item We identify visual evidence acquisition as a missing process signal in multimodal RLVR and introduce EASE, which adds visual evidence-anchored process supervision without requiring evidence metadata at inference.
    \item We map annotated evidence regions to a smoothed target distribution over visual tokens and apply this guidance only to high-reward trajectories.
    \item We validate EASE across multiple VLM backbones and benchmarks for perception-heavy reasoning, hallucination robustness, visual math, and multimodal reasoning, with diagnostics showing better alignment between visual attention and annotated evidence regions.
\end{itemize}

\begin{figure*}[t]
  \centering
  \includegraphics[width=\textwidth]{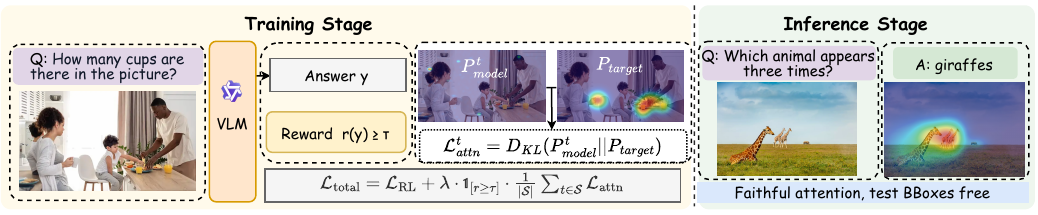}
  \caption{\textbf{Overview of EASE.} During RL training, EASE maps dataset evidence boxes to a smoothed spatial target over visual tokens. The auxiliary objective guides response-to-vision attention on high-reward trajectories toward this target, and inference uses the standard image and question inputs.}
  \label{fig:framework}
\end{figure*}

\section{Related Work}
\label{sec:related}

\textbf{RLVR for multimodal reasoning.}
RLVR has become a common VLM post-training paradigm, improving multimodal reasoning through task verifiers, rollout refinement, reflection training, chart-specific verifiable rewards, and policy optimization~\citep{yang2025r1_onevision,wang2025thinklite_vl,vl-rethinker,meng2025mm_eureka,sinha2025chartrvr,zhang2026chartr1,liu2025visual_rft,tan2026reason_rft}. However, the main training signal remains answer-level. It does not specify how the response should acquire supporting image regions.

\textbf{Visual perception signals in multimodal RL.}
Several multimodal RL methods introduce perception-oriented signals, such as clean and noisy image rollouts~\citep{liu2025noisyrollout}, perception-aware objectives~\citep{wang2025papo}, visual dependency or focus measures~\citep{huang2025vppo,wang2026vgpo}, perceptual evidence checklists~\citep{zhang2025pearl}, and grounded traces or pointing supervision~\citep{cao2025ground_r1,ni2026point_rft}. These signals encourage visual reliance through perturbations, reward-derived proxies, checklist verification, textual traces, or point-level targets, but they do not directly match response-to-vision attention to localized answer evidence. EASE instead supervises this spatial attention distribution with localized evidence regions.

\textbf{Evidence grounding and hallucination.}
Grounded VLMs connect language generation with localized visual evidence through boxes, coordinates, or grounded rationales~\citep{peng2024kosmos2,chen2023shikra,you2024ferret,xia2025bootstrapping}. Hallucination studies show that strong VLMs can produce plausible answers or reasoning traces without reliably using the image~\citep{asadi2026mirage,shi2026vlmsseeingjustsaying,wu2025gcot,xu2026morethinking,fu2025hidden}. Attention and representation analyses further show that visual information can be concentrated in middle layers, steered to reduce hallucination, or misallocated during generation~\citep{jiang2025devils,yin2025clearsight,huang2024opera,leng2024cvd,bu2025consciousgaze,xi2026lostattention}. EASE brings this grounding perspective into RL by using answer-relevant regions as training supervision for spatial attention.

\section{Method: EASE}
\label{sec:method}

Figure~\ref{fig:framework} gives an overview of EASE, which augments outcome-based multimodal RL with training-time evidence supervision. EASE converts evidence boxes into a soft visual-token target, regularizes response-to-vision attention toward this target for high-reward trajectories during actor updates.

\subsection{Motivating Observation}
We first examine whether RL with only outcome rewards still leaves failures in visual evidence acquisition. We analyze evidence-localizable examples from HallusionBench-Image with a Qwen3-VL-4B policy trained by GRPO using final-answer rewards. For each response, we compute the KL divergence between response-to-vision attention normalized over visual tokens and an evidence target constructed from annotated supporting regions. This score measures attention-target mismatch. Evidence boxes are used only to construct the diagnostic target, not fed as input to the policy.

\begin{figure}[t]
  \centering
  \includegraphics[width=\columnwidth]{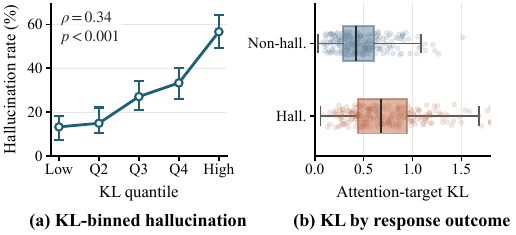}
  \caption{
  \textbf{Motivating diagnostic for outcome-only RL.}
  Higher attention-target mismatch corresponds to increased hallucination risk in (a) and a right-shifted KL distribution for hallucinated responses in (b).
  Details are in Appendix~\ref{app:motivation-diagnostic}.
  }
  \label{fig:kl-motivation}
\end{figure}

Figure~\ref{fig:kl-motivation} shows that this mismatch is statistically associated with hallucination. Figure~\ref{fig:kl-motivation}(a) shows that hallucination risk increases as KL divergence grows, and Figure~\ref{fig:kl-motivation}(b) shows that hallucinated responses have higher KL overall than responses that do not hallucinate. Together, these trends show that outcome rewards can improve final accuracy while leaving a measurable mismatch between response-to-vision attention and answer-relevant evidence. 
EASE therefore uses evidence-attention alignment as an operational process signal for visual evidence acquisition during RL.

\begin{figure*}[t]
  \centering
  \includegraphics[width=\textwidth]{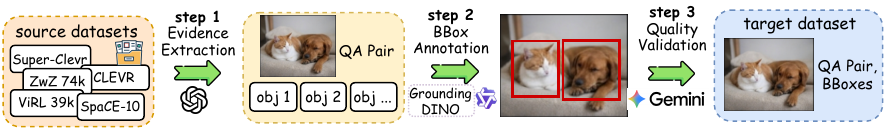}
  \caption{\textbf{Evidence annotation pipeline.} Given an image-question-answer triple, Step~1 extracts answer-relevant evidence phrases, Step~2 localizes each phrase with grounding models, and Step~3 validates the proposed boxes with semantic and geometric checks. The annotated pool contains single-evidence examples for local grounding and multi-evidence examples for cross-region reasoning. Appendix~\ref{app:pipeline} gives implementation details and dataset statistics.}
  \label{fig:pipeline}
\end{figure*}

\subsection{Evidence Annotation}
\label{sec:data}

EASE uses evidence annotations that identify the image regions supporting an answer. Each training example is represented as $(I,q,a,\mathcal{B})$, where $I$ is the original image, $q$ is the question, $a$ is the reference answer, and $\mathcal{B}=\{B_k\}_{k=1}^{K}$ is the bbox field for answer-relevant regions. These boxes are stored as training metadata and used only to construct the auxiliary attention target, leaving the image and prompt unchanged. Only training examples are annotated, and evaluation benchmarks are used without evidence metadata.

Figure~\ref{fig:pipeline} summarizes the annotation pipeline. Step~1 extracts minimal answer-relevant evidence phrases from each image-question-answer triple. Step~2 localizes each phrase with grounding models and stores the resulting coordinates as box metadata. Step~3 validates the proposed boxes with semantic and geometric checks. Existing evidence boxes are normalized and passed through the same validation stage. Appendix~\ref{app:pipeline} gives implementation details and source-level statistics.

The resulting annotated pool is split into single-evidence and multi-evidence examples for localized grounding and cross-region evidence acquisition. The default training distribution samples a balanced mixture from the two pools.

\subsection{Evidence-Anchored Attention Target}
\label{sec:target}

Let $\mathcal{V}$ denote the set of visual tokens and let $c_v$ be the image-plane center of visual token $v$, expressed in the same coordinate system as the evidence boxes. EASE converts the evidence boxes into a soft target distribution over $\mathcal{V}$. A box $B_k=(x_k,y_k,w_k,h_k)$ defines a Gaussian component centered at
\begin{equation}
\mu_k = \left(x_k + \frac{w_k}{2}, y_k + \frac{h_k}{2}\right),
\end{equation}
with diagonal covariance
\begin{equation}
\Sigma_k = \mathrm{diag}\left(\left(\frac{w_k}{4}\right)^2,\left(\frac{h_k}{4}\right)^2\right).
\end{equation}
This parameterization makes two standard deviations match half the box width and height, which concentrates most probability mass inside the annotated region while retaining tolerance to boundary noise and coarse boxes.

Rather than treating a box as a hard binary mask, we evaluate each Gaussian on the visual-token grid and normalize it into a distribution $G_k(v)$:
\begin{equation}
G_k(v)=
\frac{\mathcal{N}(c_v;\mu_k,\Sigma_k)}
{\sum_{v'\in\mathcal{V}}\mathcal{N}(c_{v'};\mu_k,\Sigma_k)}.
\end{equation}
Multi-evidence examples are represented by an equal-weight mixture:
\begin{equation}
P_{\mathrm{evid}}(v) = \frac{1}{K}\sum_{k=1}^{K}G_k(v).
\end{equation}
Equal weighting assigns the same total supervision mass to each supporting region, encouraging coverage across all evidence boxes rather than allowing large boxes to dominate the target.

We smooth the target with a uniform distribution to avoid assigning zero probability to all background tokens:
\begin{equation}
P_{\mathrm{target}}(v) = (1-\alpha)P_{\mathrm{evid}}(v) + \alpha U(v),
\label{eq:target}
\end{equation}
where $U(v)=1/|\mathcal{V}|$ and $\alpha$ controls background smoothing.

\subsection{Response-to-Vision Attention}
\label{sec:attn}

To connect evidence supervision with generation, EASE measures how response tokens attend to visual tokens during the actor update. We normalize attention logits only over the visual-token set, yielding a conditional spatial distribution over image regions. This captures where the response looks within the image without constraining its total attention mass to vision versus text. Let \(L\) denote the number of transformer layers in the language model. We extract response-to-vision attention from a semantic middle layer, setting \(\ell^\star=\lfloor 2L/3\rfloor\). This targets a stage where visual information has been semantically refined but is not yet dominated by output-adjacent lexical prediction~\citep{jiang2025devils}. For response token $t$, attention head $h$, and head dimension $d$ at layer \(\ell^\star\),
\begin{equation}
p_h^t(v)=
\frac{
\exp((\mathbf{q}_{h,t}^{\ell^\star})^{\top}\mathbf{k}_{h,v}^{\ell^\star}/\sqrt{d})
}{
\sum_{v'\in\mathcal{V}}\exp((\mathbf{q}_{h,t}^{\ell^\star})^{\top}\mathbf{k}_{h,v'}^{\ell^\star}/\sqrt{d})
}.
\end{equation}
Here, $p_h^t$ is the head-wise visual-token-normalized response-to-vision distribution. Averaging over the $H$ heads gives the model's conditional spatial attention for token $t$:
\begin{equation}
P_{\mathrm{model}}^t(v)=\frac{1}{H}\sum_{h=1}^{H}p_h^t(v).
\end{equation}
When the token belongs to trajectory $g$, we write the same distribution as $P_{\mathrm{model}}^{g,t}$.

To reduce memory overhead, we compute the attention loss on at most 64 sampled valid response tokens per sequence by default, rather than on the full generation.

\subsection{Reward-Aware RL Objective}
\label{sec:objective}

Let $\pi_\theta$ denote the VLM policy. For each prompt $x=(I,q)$, outcome-only multimodal RL samples a group of responses $\{y^{(g)}\}_{g=1}^{G}$ and obtains raw verifier rewards $r^{(g)}=R(y^{(g)},a)$. The underlying GRPO or DAPO actor loss computes group-normalized advantages from these per-response rewards. We denote this base actor objective by $\mathcal{L}_{\mathrm{RL}}$ and add EASE as an auxiliary process-supervision term. In the main experiments, we instantiate $\mathcal{L}_{\mathrm{RL}}$ with DAPO because it provides the strongest outcome-only baseline.

For a response token $t$ from trajectory $g$, the default attention loss uses model-to-target KL.
\begin{equation}
\mathcal{L}_{\mathrm{attn}}^{g,t}
=D_{\mathrm{KL}}\!\left(P_{\mathrm{model}}^{g,t} \,\|\, P_{\mathrm{target}}\right).
\label{eq:attn-kl}
\end{equation}
This direction penalizes the visual-token-normalized attention probability assigned outside the evidence-anchored target. We also evaluate the reverse direction, $D_{\mathrm{KL}}(P_{\mathrm{target}}\|P_{\mathrm{model}}^{g,t})$, as a coverage-oriented ablation.

EASE applies attention supervision only to trajectories whose raw verifier reward meets a threshold $\tau$. Let $m^{(g)}=\mathbb{1}[r^{(g)}\ge\tau]$ denote this reward gate. The total objective is
\begin{equation}
\begin{aligned}
\mathcal{L}_{\mathrm{total}}
&=
\mathcal{L}_{\mathrm{RL}}
+
\lambda_{\mathrm{attn}}\mathcal{L}_{\mathrm{EASE}},\\
\mathcal{L}_{\mathrm{EASE}}
&=
\frac{
\sum_{g=1}^{G}m^{(g)}
\frac{1}{|\mathcal{S}^{(g)}|}
\sum_{t\in\mathcal{S}^{(g)}}\mathcal{L}_{\mathrm{attn}}^{g,t}
}{
\sum_{g=1}^{G}m^{(g)}+\epsilon
}.
\end{aligned}
\label{eq:total}
\end{equation}
where $\mathcal{S}^{(g)}$ is the sampled response-token set for trajectory $g$, $\lambda_{\mathrm{attn}}$ is the attention-loss weight, and $\epsilon$ prevents division by zero when no trajectory in the group passes the gate. The gate uses raw verifier rewards rather than group-normalized advantages because it selects trajectories that can serve as reliable process demonstrations.

\subsection{Training and Inference}
\label{sec:training}

Evidence annotations are privileged training labels. They are used to construct $P_{\mathrm{target}}$ and the auxiliary actor loss during training. After training, EASE uses the same image-question pathway and decoding procedure as the underlying VLM.

\begin{table*}[t]
\centering
\caption{\textbf{Generalization across backbones.} Controlled comparison of Base, GRPO, DAPO, and EASE across three VLM backbones under the same training and evaluation protocol. Hall. denotes HallusionBench-Image. Bold and underlined scores mark the best and second-best results in each column.}
\label{tab:backbone-generalization}
\setlength{\tabcolsep}{2.4pt}
\renewcommand{\arraystretch}{1.12}
{\small
\begin{tabular}{@{}lcccccccccccc@{}}
\toprule
\multirow{2}{*}{\textbf{Method}} & \multicolumn{4}{c}{\textbf{Perception}} & \multicolumn{2}{c}{\textbf{Hallucination}} & \multicolumn{4}{c}{\textbf{Math/Geometry}} & \textbf{Logic} & \multirow{2}{*}{\textbf{Avg.}} \\
\cmidrule(lr){2-5}\cmidrule(lr){6-7}\cmidrule(lr){8-11}\cmidrule(lr){12-12}
& \textbf{HR4K} & \textbf{HR8K} & \textbf{V*} & \textbf{CV-B.} & \textbf{POPE} & \textbf{Hall.} & \textbf{MathVerse$_{V}$} & \textbf{MathVista} & \textbf{WeMath} & \textbf{MMK12} & \textbf{LogicVista} & \\
\midrule
\rowcolor{GroupHeader}
\multicolumn{13}{c}{\textit{Qwen3-VL-4B}} \\
Base & 78.3 & 72.9 & 80.1 & 85.0 & 88.5 & 68.2 & 28.0 & 73.7 & 70.7 & 58.9 & 45.3 & 68.1 \\
GRPO & 79.1 & \underline{76.3} & 86.4 & \underline{85.5} & 89.3 & 71.0 & 60.7 & \underline{75.7} & \underline{77.8} & 78.9 & 50.4 & 75.6 \\
DAPO & \underline{79.9} & 74.1 & \underline{87.4} & 85.1 & \underline{89.5} & \underline{71.3} & \underline{62.2} & 74.3 & 77.2 & \underline{80.2} & \underline{51.8} & \underline{75.7} \\
EASE (ours) & \textbf{81.3} & \textbf{79.5} & \textbf{90.1} & \textbf{87.3} & \textbf{90.8} & \textbf{74.1} & \textbf{65.1} & \textbf{77.8} & \textbf{82.9} & \textbf{81.3} & \textbf{56.9} & \textbf{78.8} \\
\addlinespace[2pt]
\rowcolor{GroupHeader}
\multicolumn{13}{c}{\textit{Qwen2.5-VL-7B}} \\
Base & 71.6 & 67.9 & 78.5 & 75.3 & 85.9 & 52.9 & 36.6 & 68.5 & 47.8 & 49.4 & 45.1 & 61.8 \\
GRPO & 74.3 & 70.3 & 83.2 & 76.9 & \underline{87.6} & 55.6 & 62.6 & \underline{70.1} & \underline{68.4} & 72.9 & 46.9 & 69.9 \\
DAPO & \underline{74.8} & \underline{70.4} & \underline{84.3} & \underline{77.0} & 87.3 & \underline{56.0} & \underline{64.5} & 68.7 & 68.0 & \underline{77.0} & \underline{47.3} & \underline{70.5} \\
EASE (ours) & \textbf{75.3} & \textbf{73.0} & \textbf{88.5} & \textbf{78.3} & \textbf{87.9} & \textbf{57.2} & \textbf{67.8} & \textbf{75.2} & \textbf{72.9} & \textbf{81.6} & \textbf{49.6} & \textbf{73.4} \\
\addlinespace[2pt]
\rowcolor{GroupHeader}
\multicolumn{13}{c}{\textit{Qwen3-VL-8B}} \\
Base & 78.9 & 74.6 & 86.4 & 85.4 & 87.0 & 71.6 & 54.6 & \underline{77.2} & 71.7 & 67.0 & 53.0 & 73.4 \\
GRPO & 80.5 & \underline{80.6} & \underline{89.0} & \underline{86.3} & 88.7 & 71.9 & 65.4 & 75.9 & 79.7 & 79.7 & 58.5 & 77.8 \\
DAPO & \underline{81.1} & 79.8 & 88.0 & 85.5 & \underline{89.1} & \underline{73.6} & \underline{65.7} & 76.0 & \underline{82.9} & \underline{80.9} & \underline{60.0} & \underline{78.4} \\
EASE (ours) & \textbf{83.8} & \textbf{81.6} & \textbf{91.6} & \textbf{87.1} & \textbf{89.6} & \textbf{77.8} & \textbf{69.7} & \textbf{78.1} & \textbf{86.4} & \textbf{82.2} & \textbf{62.3} & \textbf{80.9} \\
\bottomrule
\end{tabular}
}
\end{table*}

\section{Experiments}
\label{sec:experiments}

\subsection{Experimental Setup}

\textbf{Models and training data.}
We evaluate Qwen2.5-VL-7B~\citep{bai2025qwen25vltechnicalreport}, Qwen3-VL-4B, and Qwen3-VL-8B~\citep{bai2025qwen3}. All controlled variants use the VQA corpus annotated in Section~\ref{sec:data}, with additional statistics in Appendix~\ref{app:data}. EASE trains on a balanced mixture of examples grounded in one evidence region and examples requiring multiple evidence regions.

\textbf{Baselines.}
For controlled comparisons, we compare EASE with the supervised base model, GRPO, and DAPO under the same training data, backbone, and evaluation protocol. Table~\ref{tab:public-comparison} further places EASE in context with public 7B-scale multimodal reasoning methods, including ThinkLite-VL-7B~\citep{wang2025thinklite_vl}, VL-Rethinker-7B~\citep{vl-rethinker}, MM-Eureka-7B~\citep{meng2025mm_eureka}, NoisyRollout-7B~\citep{liu2025noisyrollout}, PAPO$_D$-7B~\citep{wang2025papo}, VPPO-RL-7B~\citep{huang2025vppo}, and VGPO-7B~\citep{wang2026vgpo}. These public rows use reported results and are intended for contextual comparison rather than strict protocol-controlled evaluation.

\textbf{Benchmarks.}
We evaluate perception with HR-Bench~\citep{hrbench}, V*~\citep{vstar}, and CV-Bench~\citep{cvbench}, hallucination with POPE~\citep{li2023pope} and HallusionBench-Image~\citep{guan2023hallusionbench}, visual math and geometry with MathVerse$_{V}$~\citep{zhang2024mathverse}, MathVista~\citep{lu2024mathvista}, WeMath~\citep{qiao2024wemath}, and MMK12~\citep{meng2025mm_eureka}, and broader multimodal reasoning with LogicVista~\citep{xiao2024LogicVista}. MathVerse$_{V}$ denotes the vision-centric subset of MathVerse.

\textbf{Implementation details.}
Unless otherwise noted, models are trained for 2 epochs with learning rate \(1\times10^{-6}\). EASE uses the smoothed Gaussian-mixture target with \(\alpha=0.1\), model-to-target KL, \(\lambda_{\mathrm{attn}}=0.001\), reward-1 gating under the 0/1 verifier, and at most 64 valid response tokens for the attention loss. Additional implementation details are provided in Appendix~\ref{app:implementation}.

\begin{table*}[t]
\centering
\caption{\textbf{Contextual comparison with prior 7B-scale multimodal reasoning methods.} Prior scores are taken from the VGPO report, where available. "\(^{*}\)" indicates that NoisyRollout is trained on Geo3K. Bold and underlined scores indicate the best and second-best results in each column.}
\label{tab:public-comparison}
{\small
\begin{tabular*}{0.92\textwidth}{@{\extracolsep{\fill}}lccccccc@{}}
\toprule
\textbf{Method} & \textbf{MathVista} & \textbf{MathVerse$_{V}$} & \textbf{WeMath} & \textbf{MMK12} & \textbf{Geo3K} & \textbf{LogicVista} & \textbf{Avg.} \\
\midrule
ThinkLite-VL-7B & 73.6 & 30.9 & 43.7 & 50.5 & 45.3 & 44.3 & 48.1 \\
VL-Rethinker-7B & 65.2 & 64.8 & 67.5 & 66.0 & 44.4 & 47.0 & 59.2 \\
MM-Eureka-7B & 66.3 & 63.6 & 66.6 & 61.7 & 41.1 & 47.9 & 57.9 \\
NoisyRollout-7B & 72.9 & 51.7 & 64.4 & 51.2 & \textbf{55.2}\(^{*}\) & 47.0 & 57.1 \\
PAPO$_D$-7B & 72.3 & 66.6 & 69.4 & 80.8 & 48.8 & 45.9 & 64.0 \\
VPPO-RL-7B & 70.5 & \underline{67.6} & 70.6 & \textbf{81.8} & 46.9 & 48.8 & 64.4 \\
VGPO-7B & \underline{74.1} & \underline{67.6} & \underline{72.5} & 81.5 & 45.8 & \underline{49.4} & \underline{65.2} \\
\addlinespace[2pt]
EASE-7B (ours) & \textbf{75.2} & \textbf{67.8} & \textbf{72.9} & \underline{81.6} & \underline{48.9} & \textbf{49.6} & \textbf{66.0} \\
\bottomrule
\end{tabular*}
}
\end{table*}

\begin{table}[t]
\centering
\caption{\textbf{Core ablations on Qwen3-VL-4B.} We ablate reward gating, evidence-target construction and validity, evidence composition, and the attention extraction layer. \(L\) denotes the number of language-model blocks, and the default EASE layer is \(\ell^\star=\lfloor 2L/3\rfloor\).}
\label{tab:key-ablations}
\setlength{\tabcolsep}{3pt}
\renewcommand{\arraystretch}{1.12}
\resizebox{\columnwidth}{!}{%
\begin{tabular}{@{}lcccc@{}}
\toprule
\textbf{Variant} & \textbf{V*} & \textbf{Hall.} & \textbf{MMK12} & \textbf{Avg.} \\
\midrule
EASE (ours) & \textbf{90.1} & \textbf{74.1} & \textbf{81.3} & \textbf{81.8} \\
\addlinespace[2pt]
\rowcolor{GroupHeader}
\multicolumn{5}{c}{\textit{Reward-aware supervision}} \\
w/o reward gating & 86.4 & 71.0 & 77.9 & 78.4 \\
\addlinespace[2pt]
\rowcolor{GroupHeader}
\multicolumn{5}{c}{\textit{Evidence target construction}} \\
hard box target & 87.8 & 71.6 & 78.5 & 79.3 \\
target-to-model KL & 89.5 & 72.8 & 80.6 & 81.0 \\
w/o smoothing & 88.2 & 72.0 & 79.3 & 79.8 \\
\addlinespace[2pt]
\rowcolor{GroupHeader}
\multicolumn{5}{c}{\textit{Evidence target validity and robustness}} \\
50\% box jitter (IoU $\approx 0.7$) & 89.3 & 73.5 & 80.7 & 81.2 \\
random targets & 81.9 & 69.3 & 77.5 & 76.2 \\
non-overlapping targets & 82.2 & 68.5 & 78.2 & 76.3 \\
\addlinespace[2pt]
\rowcolor{GroupHeader}
\multicolumn{5}{c}{\textit{Evidence composition}} \\
single-only & 89.5 & 72.7 & 78.9 & 80.4 \\
multi-only & 88.0 & 73.2 & 80.7 & 80.6 \\
\addlinespace[2pt]
\rowcolor{GroupHeader}
\multicolumn{5}{c}{\textit{Attention extraction granularity}} \\
early layer (\(L/4\)) & 86.8 & 71.3 & 77.8 & 78.6 \\
central middle layer (\(L/2\)) & 88.7 & 72.8 & 79.6 & 80.4 \\
final layer (\(L\)) & 86.2 & 70.8 & 77.2 & 78.1 \\
\bottomrule
\end{tabular}%
}
\end{table}

\subsection{Main Results}
\label{sec:main-results}

\textbf{Consistent Gains under Controlled Training.}
The controlled comparison in Table~\ref{tab:backbone-generalization} more directly isolates the effect of the proposed training signal, since Base, GRPO, DAPO, and EASE are evaluated under the same protocol within each backbone. On Qwen2.5-VL-7B, EASE improves DAPO on every reported benchmark and raises the average from 70.5 to 73.4. The largest gains appear on vision-centric mathematical reasoning and fine-grained perception, including MathVista (+6.5), WeMath (+4.9), MMK12 (+4.6), and V* (+4.2). This pattern aligns with the purpose of EASE, where the auxiliary loss serves as a process signal that encourages the policy's response-to-vision attention to concentrate on answer-relevant visual evidence before generating the response.

\textbf{Scalability across Model Scales and Families.}
The same trend is visible across Qwen3-VL backbones. For Qwen3-VL-4B, EASE improves the DAPO average from 75.7 to 78.8 and gives gains across the full benchmark suite, including fine-grained perception, hallucination, math/geometry, and LogicVista. For Qwen3-VL-8B, EASE reaches the best average score among the compared variants, improving DAPO from 78.4 to 80.9 and again improving every reported benchmark. These results suggest that evidence-focused attention regularization transfers across model generations and scales when applied under the same training and evaluation protocol.

\textbf{Contextual Comparison with Public RL Methods.}
Table~\ref{tab:public-comparison} situates EASE among recent 7B-scale multimodal reasoning methods on shared benchmarks from prior work. EASE-7B achieves the best average score in this comparison and obtains the strongest results on MathVista, MathVerse$_V$, WeMath, and LogicVista. It also ranks second on MMK12, showing that evidence-anchored attention supervision remains effective across several forms of vision-centric reasoning. These results indicate that EASE is competitive with leading 7B-scale multimodal RL methods.

\subsection{Ablation Studies}
\label{sec:ablations}

\textbf{Ablation study on reward-aware supervision.}
We ablate the reward gate that restricts attention supervision to trajectories with positive outcome rewards. Removing this gate applies the attention loss to all sampled responses and causes consistent drops across the representative benchmarks in Table~\ref{tab:key-ablations}. This suggests that evidence boxes alone are insufficient when the generated response is wrong, since low-reward rollouts can provide misleading response-to-evidence alignments. Reward-aware gating therefore helps EASE treat only successful trajectories as reliable process demonstrations.

\begin{figure}[H]
\centering
\includegraphics[width=\columnwidth]{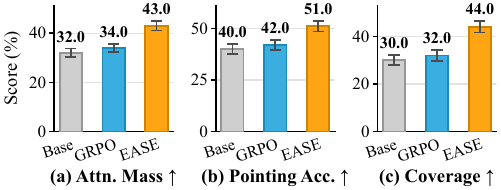}
\caption{\textbf{Evidence-acquisition diagnostics.}
On Qwen3-VL-4B, EASE increases response-to-vision attention on annotated evidence regions compared with Base and GRPO, as measured by attention mass, pointing accuracy, and multi-evidence coverage on a held-out validation set. Error bars denote 95\% CIs.}
\label{fig:attention-faithfulness}
\end{figure}

\begin{figure}[H]
\centering
\includegraphics[width=\columnwidth]{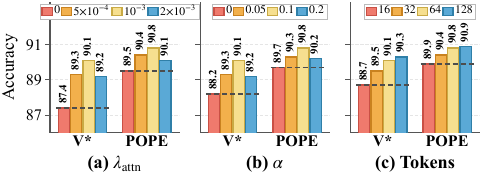}
\caption{\textbf{Hyperparameter sensitivity.} On the Qwen3-VL-4B backbone, we vary the attention-loss weight, background smoothing, and sampled response-token budget around the default EASE configuration. Marked bars indicate default settings.}
\label{fig:hyperparam-sensitivity}
\end{figure}

\begin{figure*}[t]
\centering
\includegraphics[width=\textwidth]{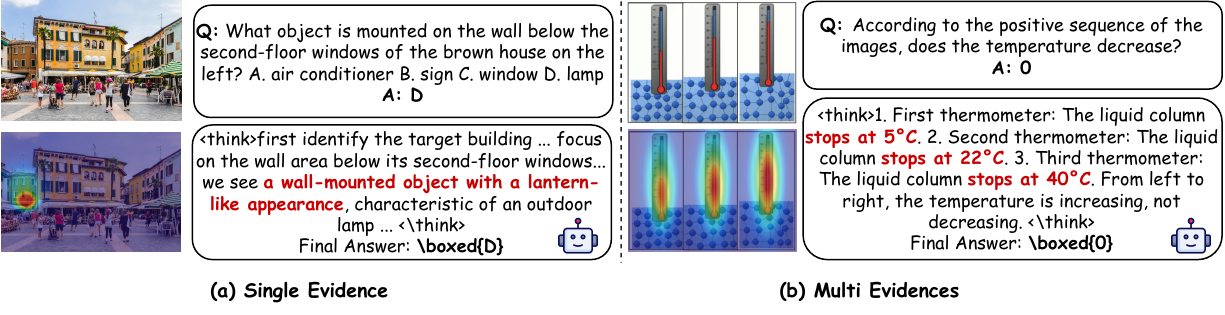}
\caption{\textbf{Qualitative evidence analysis.} Representative examples with EASE reasoning traces, final answers, and response-to-vision attention maps. Attention maps are used only for analysis.}
\label{fig:attention-cases}
\end{figure*}

\textbf{Ablation study on evidence target construction.}
We ablate three components of the evidence target, including Gaussian smoothing, background smoothing, and the KL direction. Replacing the Gaussian target with a hard box mask causes the largest drop, suggesting that binary supervision is brittle when evidence spans object parts, coarse boxes, or multiple visual tokens. Removing background smoothing also hurts performance, indicating that a small amount of background mass helps avoid over-constraining contextual visual tokens. Reversing the KL direction performs closer to the default but remains weaker. These results support the smoothed Gaussian target with KL from model attention to the evidence target, which guides attention toward evidence regions while preserving tolerance to annotation noise and useful context.

\textbf{Ablation study on evidence target validity and robustness.}
We evaluate target validity and robustness by jittering boxes in 50\% of training examples to an IoU of approximately 0.7, replacing targets with randomly placed boxes of the same size, and translating targets to non-overlapping regions. As shown in Table~\ref{tab:key-ablations}, moderate jitter causes only a 0.6-point average decrease and EASE remains 1.6 points above DAPO, whereas random and non-overlapping targets both fall below DAPO. These results show that EASE tolerates moderate localization error but requires approximately correct spatial supervision.

\textbf{Ablation study on evidence composition.}
We compare training on one-region examples, multi-region examples, and their balanced mixture. The one-region setting performs well on V*, which emphasizes fine-grained localization, but drops more on MMK12, where many examples require multiple visual cues. Conversely, the multi-region setting better preserves MMK12 but loses more on V*. The balanced mixture gives the best overall result by combining local grounding with cross-region evidence acquisition.

\textbf{Ablation study on attention extraction granularity.}
We ablate the layer used to extract response-to-vision attention. Early-layer supervision performs poorly, suggesting that shallow visual representations are not yet sufficiently aligned with answer-level evidence. The central middle layer improves over the early layer but remains below the default setting. The final layer also degrades performance, likely because late attention is more affected by lexical prediction and response formatting than by evidence acquisition. These results support extracting attention from \(\ell^\star=\lfloor 2L/3\rfloor\), where visual tokens have been semantically refined while still retaining useful spatial evidence signals.

\subsection{Further Analysis}
\label{sec:further-analysis}

\textbf{Evidence-acquisition diagnostics.}
We measure response-to-vision attention on a held-out evidence-annotated validation set, using evidence boxes only for offline analysis. Figure~\ref{fig:attention-faithfulness} reports attention mass, pointing accuracy, and multi-evidence coverage, which respectively measure attention inside evidence regions, whether the strongest attention peak lands on evidence, and whether multiple supporting regions are covered. EASE improves all three metrics over both the base model and outcome-only GRPO, suggesting that the auxiliary objective improves attention-based evidence-acquisition diagnostics rather than only final-answer accuracy. Detailed metric definitions are provided in Appendix~\ref{app:attention-metrics}.

\begin{table}[t]
\centering
\caption{\textbf{Evidence interventions.} Accuracy on 1,000 held-out evidence-annotated examples after masking annotated evidence or suppressing evidence-token attention. The matched control masks an area- and shape-matched non-evidence region or suppresses equal attention mass from non-evidence tokens. Question-only applies to masking. Gap is matched-control accuracy minus evidence-intervention accuracy.}
\label{tab:evidence-interventions}
\small
\setlength{\tabcolsep}{0.8pt}
\renewcommand{\arraystretch}{1.08}
\begin{tabular}{@{}lccccc@{}}
\toprule
\textbf{Method} & \tablehead{Original} & \tableheadtwo{Evidence}{intervention} & \tableheadtwo{Matched}{control} & \tableheadtwo{Question}{only} & \textbf{Gap} \\
\midrule
\rowcolor{GroupHeader}
\multicolumn{6}{c}{\textit{Masking}} \\
DAPO & 79.0 & 19.6 & 73.4 & 12.8 & 53.8 \\
EASE (ours) & 82.3 & 10.9 & 76.1 & 12.5 & 65.2 \\
\addlinespace[1pt]
\rowcolor{GroupHeader}
\multicolumn{6}{c}{\textit{Attention suppression}} \\
DAPO & 79.0 & 74.8 & 77.6 & -- & 2.8 \\
EASE (ours) & 82.3 & 74.4 & 80.5 & -- & 6.1 \\
\bottomrule
\end{tabular}
\end{table}

\textbf{Functional validation of evidence use.}
Attention alignment alone does not establish whether the model functionally relies on the aligned visual evidence. On 1,000 held-out examples, we mask annotated evidence or an area- and shape-matched non-evidence region. We also suppress response-token attention to evidence tokens at the supervised layer and redistribute the removed mass over the remaining visual tokens, with a matched control that suppresses equal mass from non-evidence tokens. Table~\ref{tab:evidence-interventions} shows larger evidence-specific gaps for EASE than DAPO under both interventions. Evidence masking also produces more correct-to-incorrect flips for EASE, whereas matched masking has little effect. These results support stronger functional reliance on annotated evidence, while not treating the supervised attention layer as a complete causal account of visual information use.

\textbf{Hyperparameter sensitivity.}
We further test whether EASE relies on narrow hyperparameter choices. Figure~\ref{fig:hyperparam-sensitivity} varies the attention-loss weight \(\lambda_{\mathrm{attn}}\), background smoothing \(\alpha\), and the sampled token budget around the default configuration. Setting \(\lambda_{\mathrm{attn}}=0\) recovers outcome-only DAPO, while moderate attention supervision improves both V* and POPE. Very large weights reduce accuracy, indicating that the auxiliary loss should guide rather than dominate policy optimization. A small amount of background smoothing is also beneficial, with \(\alpha=0.1\) giving the best or near-best scores. Finally, sampling 64 response tokens matches or nearly matches the 128-token setting, supporting the memory-efficient default.

\textbf{Qualitative evidence analysis.}
Figure~\ref{fig:attention-cases} provides qualitative examples of how EASE changes response-to-vision attention during generation. In these cases, the model produces correct reasoning traces and final answers while its attention concentrates on regions that visually support the answer. These examples illustrate the intended effect of the auxiliary objective, which encourages high-reward trajectories to align response tokens with answer-relevant evidence regions. Appendix~\ref{app:qualitative} further analyzes representative failure cases to clarify the boundary of this effect.

The quantitative reasoning-quality evaluation is reported in Appendix~\ref{app:reasoning-quality}.

\section{Conclusion}

We presented EASE, a framework that adds evidence-anchored process supervision to multimodal RLVR. EASE converts annotated evidence regions into soft spatial targets and regularizes response-to-vision attention for high-reward trajectories, encouraging successful responses to align with visually grounded evidence. Across three VLM backbones and diverse reasoning and hallucination benchmarks, EASE improves outcome-only RL baselines and strengthens measured evidence acquisition. These results suggest that supervising visual evidence acquisition is a practical complement to final-answer rewards for more grounded multimodal reasoning.


\section*{Limitations}

EASE relies on evidence annotations during training. Although these annotations are used only for the auxiliary loss and are not provided at inference, obtaining reliable boxes introduces additional data construction cost. The method is also most suitable for tasks whose supporting evidence can be localized to one or more image regions, and box-based supervision may be less suitable for holistic, stylistic, affective, or globally distributed cues.
Finally, EASE mainly supervises where response-to-vision attention is allocated within the image, rather than directly optimizing how much the generation depends on visual input. We therefore interpret the attention diagnostics as evidence of improved spatial grounding, not as a complete causal measure of visual reliance.

\section*{Acknowledgments}
This work is supported by the Zhongguancun Academy, (Grant No.s C20250203).

\section*{Ethical Considerations}

This work is intended for research on multimodal RLVR and grounded reasoning, not for direct deployment in safety-critical settings. Additional details on artifact use, data characteristics, model-assisted annotation, computation, and AI-assistant use are provided in Appendix~\ref{app:responsible_details}.

\bibliography{custom}

@inproceedings{liu2024groundingdino,
  title={Grounding dino: Marrying dino with grounded pre-training for open-set object detection},
  author={Shilong Liu and Zhaoyang Zeng and Tianhe Ren and Feng Li and Hao Zhang and Jie Yang and Qing Jiang and Chunyuan Li and Jianwei Yang and Hang Su and Jun Zhu and Lei Zhang},
  booktitle={European conference on computer vision},
  pages={38--55},
  year={2024},
  organization={Springer}
}

@inproceedings{data_clevr,
  title={Clevr: A diagnostic dataset for compositional language and elementary visual reasoning},
  author={Johnson, Justin and Hariharan, Bharath and Van Der Maaten, Laurens and Fei-Fei, Li and Lawrence Zitnick, C and Girshick, Ross},
  booktitle={Proceedings of the IEEE conference on computer vision and pattern recognition},
  pages={2901--2910},
  year={2017}
}

@inproceedings{data_super_clevr,
  title={Super-clevr: A virtual benchmark to diagnose domain robustness in visual reasoning},
  author={Li, Zhuowan and Wang, Xingrui and Stengel-Eskin, Elias and Kortylewski, Adam and Ma, Wufei and Van Durme, Benjamin and Yuille, Alan L},
  booktitle={Proceedings of the IEEE/CVF conference on computer vision and pattern recognition},
  pages={14963--14973},
  year={2023}
}

@misc{data_space10,
      title={SpaCE-10: A Comprehensive Benchmark for Multimodal Large Language Models in Compositional Spatial Intelligence}, 
      author={Ziyang Gong and Wenhao Li and Oliver Ma and Songyuan Li and Zhaokai Wang and Songyuan Li and Jiayi Ji and Xue Yang and Gen Luo and Junchi Yan and Rongrong Ji},
      year={2025},
      eprint={2506.07966},
      archivePrefix={arXiv},
      primaryClass={cs.CV},
      url={https://arxiv.org/abs/2506.07966}, 
}

@article{wei2026zwz,
  title={Zooming without Zooming: Region-to-Image Distillation for Fine-Grained Multimodal Perception},
  author={Lai Wei and Liangbo He and Jun Lan and Lingzhong Dong and Yutong Cai and Siyuan Li and Huijia Zhu and Weiqiang Wang and Linghe Kong and Yue Wang and Zhuosheng Zhang and Weiran Huang},
  journal={arXiv preprint arXiv:2602.11858},
  year={2026}
}

@inproceedings{jiang2025devils,
  title={Devils in middle layers of large vision-language models: Interpreting, detecting and mitigating object hallucinations via attention lens},
  author={Jiang, Zhangqi and Chen, Junkai and Zhu, Beier and Luo, Tingjin and Shen, Yankun and Yang, Xu},
  booktitle={Proceedings of the IEEE/CVF Conference on Computer Vision and Pattern Recognition},
  pages={25004--25014},
  year={2025}
}

@article{grpo,
  title={Deepseekmath: Pushing the limits of mathematical reasoning in open language models},
  author={Zhihong Shao and Peiyi Wang and Qihao Zhu and Runxin Xu and Junxiao Song and Xiao Bi and Haowei Zhang and Mingchuan Zhang and Y. K. Li and Y. Wu and Daya Guo},
  journal={arXiv preprint arXiv:2402.03300},
  year={2024}
}

@article{dapo,
  title={Dapo: An open-source llm reinforcement learning system at scale},
  author={Yu, Qiying and Zhang, Zheng and Zhu, Ruofei and Yuan, Yufeng and Zuo, Xiaochen and Yue, Yu and Dai, Weinan and Fan, Tiantian and Liu, Gaohong and Liu, Lingjun and others},
  journal={Advances in Neural Information Processing Systems},
  volume={38},
  pages={113222--113244},
  year={2026}
}

@article{wang2026vgpo,
  title={Visually-Guided Policy Optimization for Multimodal Reasoning},
  author={Wang, Zengbin and Xiong, Feng and Lin, Liang and Hu, Xuecai and Wang, Yong and Wang, Yanlin and Zhang, Man and Chu, Xiangxiang},
  journal={arXiv preprint arXiv:2604.09349},
  year={2026}
}

@misc{bai2025qwen3,
      title={Qwen3-VL Technical Report}, 
      author={Shuai Bai and Yuxuan Cai and Ruizhe Chen and Keqin Chen and Xionghui Chen and Zesen Cheng and Lianghao Deng and Wei Ding and Chang Gao and Chunjiang Ge and Wenbin Ge and Zhifang Guo and Qidong Huang and Jie Huang and Fei Huang and Binyuan Hui and Shutong Jiang and Zhaohai Li and Mingsheng Li and Mei Li and Kaixin Li and Zicheng Lin and Junyang Lin and Xuejing Liu and Jiawei Liu and Chenglong Liu and Yang Liu and Dayiheng Liu and Shixuan Liu and Dunjie Lu and Ruilin Luo and Chenxu Lv and Rui Men and Lingchen Meng and Xuancheng Ren and Xingzhang Ren and Sibo Song and Yuchong Sun and Jun Tang and Jianhong Tu and Jianqiang Wan and Peng Wang and Pengfei Wang and Qiuyue Wang and Yuxuan Wang and Tianbao Xie and Yiheng Xu and Haiyang Xu and Jin Xu and Zhibo Yang and Mingkun Yang and Jianxin Yang and An Yang and Bowen Yu and Fei Zhang and Hang Zhang and Xi Zhang and Bo Zheng and Humen Zhong and Jingren Zhou and Fan Zhou and Jing Zhou and Yuanzhi Zhu and Ke Zhu},
      year={2025},
      eprint={2511.21631},
      archivePrefix={arXiv},
      primaryClass={cs.CV},
      url={https://arxiv.org/abs/2511.21631}, 
}

@misc{bai2025qwen25vltechnicalreport,
      title={Qwen2.5-VL Technical Report}, 
      author={Shuai Bai and Keqin Chen and Xuejing Liu and Jialin Wang and Wenbin Ge and Sibo Song and Kai Dang and Peng Wang and Shijie Wang and Jun Tang and Humen Zhong and Yuanzhi Zhu and Mingkun Yang and Zhaohai Li and Jianqiang Wan and Pengfei Wang and Wei Ding and Zheren Fu and Yiheng Xu and Jiabo Ye and Xi Zhang and Tianbao Xie and Zesen Cheng and Hang Zhang and Zhibo Yang and Haiyang Xu and Junyang Lin},
      year={2025},
      eprint={2502.13923},
      archivePrefix={arXiv},
      primaryClass={cs.CV},
      url={https://arxiv.org/abs/2502.13923},
}

@article{wang2025thinklite_vl,
  title={SoTA with Less: MCTS-Guided Sample Selection for Data-Efficient Visual Reasoning Self-Improvement},
  author={Wang, Xiyao and Yang, Zhengyuan and Feng, Chao and Lu, Hongjin and Li, Linjie and Lin, Chung-Ching and Lin, Kevin and Huang, Furong and Wang, Lijuan},
  journal={arXiv preprint arXiv:2504.07934},
  year={2025}
}

@article{vl-rethinker,
  title={VL-Rethinker: Incentivizing Self-Reflection of Vision-Language Models with Reinforcement Learning},
  author = {Wang, Haozhe and Qu, Chao and Huang, Zuming and Chu, Wei and Lin, Fangzhen and Chen, Wenhu},
  journal={arXiv preprint arXiv:2504.08837},
  year={2025}
}

@article{meng2025mm_eureka,
  title={Mm-eureka: Exploring the frontiers of multimodal reasoning with rule-based reinforcement learning},
  author={Fanqing Meng and Lingxiao Du and Zongkai Liu and Zhixiang Zhou and Quanfeng Lu and Daocheng Fu and Tiancheng Han and Botian Shi and Wenhai Wang and Junjun He and Kaipeng Zhang and Ping Luo and Yu Qiao and Qiaosheng Zhang and Wenqi Shao},
  journal={arXiv preprint arXiv:2503.07365},
  year={2025}
}

@article{liu2025noisyrollout,
  title={Noisyrollout: Reinforcing visual reasoning with data augmentation},
  author={Liu, Xiangyan and Ni, Jinjie and Wu, Zijian and Du, Chao and Dou, Longxu and Wang, Haonan and Pang, Tianyu and Shieh, Michael Qizhe},
  journal={arXiv preprint arXiv:2504.13055},
  year={2025}
}

@article{wang2025papo,
  title={Perception-Aware Policy Optimization for Multimodal Reasoning},
  author={Zhenhailong Wang and Xuehang Guo and Sofia Stoica and Haiyang Xu and Hongru Wang and Hyeonjeong Ha and Xiusi Chen and Yangyi Chen and Ming Yan and Fei Huang and Heng Ji},
  journal={arXiv preprint arXiv:2507.06448},
  year={2025}
}

@article{huang2025vppo,
  title={Spotlight on token perception for multimodal reinforcement learning},
  author={Huang, Siyuan and Qu, Xiaoye and Li, Yafu and Luo, Yun and He, Zefeng and Liu, Daizong and Cheng, Yu},
  journal={arXiv preprint arXiv:2510.09285},
  year={2025}
}

@inproceedings{hrbench,
  title={Divide, conquer and combine: A training-free framework for high-resolution image perception in multimodal large language models},
  author={Wang, Wenbin and Ding, Liang and Zeng, Minyan and Zhou, Xiabin and Shen, Li and Luo, Yong and Yu, Wei and Tao, Dacheng},
  booktitle={Proceedings of the AAAI Conference on Artificial Intelligence},
  volume={39},
  number={8},
  pages={7907--7915},
  year={2025}
}

@inproceedings{vstar,
  title={V*: Guided visual search as a core mechanism in multimodal llms},
  author={Wu, Penghao and Xie, Saining},
  booktitle={Proceedings of the IEEE/CVF Conference on Computer Vision and Pattern Recognition},
  pages={13084--13094},
  year={2024}
}

@misc{cvbench,
  title={Cambrian-1: A Fully Open, Vision-Centric Exploration of Multimodal LLMs},
  author={Shengbang Tong and Ellis Brown and Penghao Wu and Sanghyun Woo and Manoj Middepogu and Sai Charitha Akula and Jihan Yang and Shusheng Yang and Adithya Iyer and Xichen Pan and Austin Wang and Rob Fergus and Yann LeCun and Saining Xie},
  year={2024},
  eprint={2406.16860},
}

@inproceedings{lu2024mathvista,
  title={MathVista: Evaluating Mathematical Reasoning of Foundation Models in Visual Contexts},
  author={Lu, Pan and Bansal, Hritik and Xia, Tony and Liu, Jiacheng and Li, Chunyuan and Hajishirzi, Hannaneh and Cheng, Hao and Chang, Kai-Wei and Galley, Michel and Gao, Jianfeng},
  booktitle={International Conference on Learning Representations (ICLR)},
  year={2024}
}

@article{zhang2024mathverse,
  title={MathVerse: Does Your Multi-modal LLM Truly See the Diagrams in Visual Math Problems?},
  author={Renrui Zhang and Dongzhi Jiang and Yichi Zhang and Haokun Lin and Ziyu Guo and Pengshuo Qiu and Aojun Zhou and Pan Lu and Kai-Wei Chang and Peng Gao and Hongsheng Li},
  journal={arXiv preprint arXiv:2403.14624},
  year={2024}
}

@article{qiao2024wemath,
  title={We-Math: Does Your Large Multimodal Model Achieve Human-like Mathematical Reasoning?},
  author={Runqi Qiao and Qiuna Tan and Guanting Dong and Minhui Wu and Chong Sun and Xiaoshuai Song and Zhuoma GongQue and Shanglin Lei and Zhe Wei and Miaoxuan Zhang and Runfeng Qiao and Yifan Zhang and Xiao Zong and Yida Xu and Muxi Diao and Zhimin Bao and Chen Li and Honggang Zhang},
  journal={arXiv preprint arXiv:2407.01284},
  year={2024}
}

@misc{xiao2024LogicVista,
  title={LogicVista: Multimodal LLM Logical Reasoning Benchmark in Visual Contexts}, 
  author={Yijia Xiao and Edward Sun and Tianyu Liu and Wei Wang},
  year={2024},
  eprint={2407.04973},
  archivePrefix={arXiv},
  primaryClass={cs.AI},
  url={https://arxiv.org/abs/2407.04973}, 
}

@article{li2023pope,
  title={Evaluating object hallucination in large vision-language models},
  author={Li, Yifan and Du, Yifan and Zhou, Kun and Wang, Jinpeng and Zhao, Wayne Xin and Wen, Ji-Rong},
  journal={arXiv preprint arXiv:2305.10355},
  year={2023}
}

@misc{guan2023hallusionbench,
  title={HallusionBench: An Advanced Diagnostic Suite for Entangled Language Hallucination \& Visual Illusion in Large Vision-Language Models}, 
  author={Tianrui Guan and Fuxiao Liu and Xiyang Wu and Ruiqi Xian and Zongxia Li and Xiaoyu Liu and Xijun Wang and Lichang Chen and Furong Huang and Yaser Yacoob and Dinesh Manocha and Tianyi Zhou},
  year={2023},
  eprint={2310.14566},
  archivePrefix={arXiv},
  primaryClass={cs.CV}
}

@inproceedings{yang2025r1_onevision,
  title={R1-onevision: Advancing generalized multimodal reasoning through cross-modal formalization},
  author={Yi Yang and Xiaoxuan He and Hongkun Pan and Xiyan Jiang and Yan Deng and Xingtao Yang and Haoyu Lu and Dacheng Yin and Fengyun Rao and Minfeng Zhu and Bo Zhang and Wei Chen},
  booktitle={Proceedings of the IEEE/CVF International Conference on Computer Vision},
  pages={2376--2385},
  year={2025}
}

@article{asadi2026mirage,
  title={Mirage: The illusion of visual understanding},
  author={Asadi, Mohammad and O'Sullivan, Jack W and Cao, Fang and Nedaee, Tahoura and Rajabalifardi, Kamyar and Li, Fei-Fei and Adeli, Ehsan and Ashley, Euan},
  journal={arXiv preprint arXiv:2603.21687},
  year={2026}
}

@misc{shi2026vlmsseeingjustsaying,
      title={Are VLMs Seeing or Just Saying? Uncovering the Illusion of Visual Re-examination}, 
      author={Chufan Shi and Cheng Yang and Yaokang Wu and Linhao Jin and Bo Shui and Taylor Berg-Kirkpatrick and Xuezhe Ma},
      year={2026},
      eprint={2605.15864},
      archivePrefix={arXiv},
      primaryClass={cs.CV},
      url={https://arxiv.org/abs/2605.15864}, 
}

@article{wu2025gcot,
  title={Grounded chain-of-thought for multimodal large language models},
  author={Wu, Qiong and Yang, Xiangcong and Zhou, Yiyi and Fang, Chenxin and Song, Baiyang and Sun, Xiaoshuai and Ji, Rongrong},
  journal={arXiv preprint arXiv:2503.12799},
  year={2025}
}

@inproceedings{yin2025clearsight,
  title={Clearsight: Visual signal enhancement for object hallucination mitigation in multimodal large language models},
  author={Yin, Hao and Si, Guangzong and Wang, Zilei},
  booktitle={Proceedings of the Computer Vision and Pattern Recognition Conference},
  pages={14625--14634},
  year={2025}
}

@article{xi2026lostattention,
  title={Large Vision-Language Models Get Lost in Attention},
  author={Xi, Gongli and Tian, Ye and Yang, Mengyu and Yi, Huahui and Lin, Liang and Hao, Xiaoshuai and Wang, Kun and Wang, Wendong},
  journal={arXiv preprint arXiv:2605.05668},
  year={2026}
}

@inproceedings{liu2025visual_rft,
  title={Visual-rft: Visual reinforcement fine-tuning},
  author={Liu, Ziyu and Sun, Zeyi and Zang, Yuhang and Dong, Xiaoyi and Cao, Yuhang and Duan, Haodong and Lin, Dahua and Wang, Jiaqi},
  booktitle={Proceedings of the IEEE/CVF International Conference on Computer Vision},
  pages={2034--2044},
  year={2025}
}

@article{tan2026reason_rft,
  title={Reason-rft: Reinforcement fine-tuning for visual reasoning of vision language models},
  author={Tan, Huajie and Ji, Yuheng and Hao, Xiaoshuai and Chen, Xiansheng and Wang, Pengwei and Wang, Zhongyuan and Zhang, Shanghang},
  journal={Advances in neural information processing systems},
  volume={38},
  pages={5772--5822},
  year={2026}
}

@article{cao2025ground_r1,
  title={Ground-r1: Incentivizing grounded visual reasoning via reinforcement learning},
  author={Cao, Meng and Zhao, Haoze and Zhang, Can and Chang, Xiaojun and Reid, Ian and Liang, Xiaodan},
  journal={arXiv preprint arXiv:2505.20272},
  year={2025}
}

@article{ni2026point_rft,
  title={Point-rft: Improving multimodal reasoning with visually grounded reinforcement finetuning},
  author={Ni, Minheng and Yang, Zhengyuan and Li, Linjie and Lin, Chung-Ching and Lin, Kevin and Zuo, Wangmeng and Wang, Lijuan},
  journal={Advances in Neural Information Processing Systems},
  volume={38},
  pages={20538--20559},
  year={2026}
}

@article{xu2026morethinking,
  title={More thinking, less seeing? assessing amplified hallucination in multimodal reasoning models},
  author={Xu, Zhongxing and Liu, Chengzhi and Wei, Qingyue and Wu, Juncheng and Zou, James and Wang, Xin and Zhou, Yuyin and Liu, Sheng},
  journal={Advances in Neural Information Processing Systems},
  volume={38},
  pages={82878--82905},
  year={2026}
}

@article{fu2025hidden,
  title={Hidden in plain sight: Vlms overlook their visual representations},
  author={Fu, Stephanie and Bonnen, Tyler and Guillory, Devin and Darrell, Trevor},
  journal={arXiv preprint arXiv:2506.08008},
  year={2025}
}

@inproceedings{huang2024opera,
  title={Opera: Alleviating hallucination in multi-modal large language models via over-trust penalty and retrospection-allocation},
  author={Huang, Qidong and Dong, Xiaoyi and Zhang, Pan and Wang, Bin and He, Conghui and Wang, Jiaqi and Lin, Dahua and Zhang, Weiming and Yu, Nenghai},
  booktitle={Proceedings of the IEEE/CVF Conference on Computer Vision and Pattern Recognition},
  pages={13418--13427},
  year={2024}
}

@inproceedings{leng2024cvd,
  title={Mitigating object hallucinations in large vision-language models through visual contrastive decoding},
  author={Leng, Sicong and Zhang, Hang and Chen, Guanzheng and Li, Xin and Lu, Shijian and Miao, Chunyan and Bing, Lidong},
  booktitle={Proceedings of the IEEE/CVF Conference on Computer Vision and Pattern Recognition},
  pages={13872--13882},
  year={2024}
}

@article{bu2025consciousgaze,
  title={Conscious Gaze: Adaptive Attention Mechanisms for Hallucination Mitigation in Vision-Language Models},
  author={Bu, Weijue and Yuan, Guan and Zhang, Guixian},
  journal={arXiv preprint arXiv:2512.05546},
  year={2025}
}

@inproceedings{peng2024kosmos2,
  title={Grounding multimodal large language models to the world},
  author={Peng, Zhiliang and Wang, Wenhui and Dong, Li and Hao, Yaru and Huang, Shaohan and Ma, Shuming and Ye, Qixiang and Wei, Furu},
  booktitle={International Conference on Learning Representations},
  volume={2024},
  pages={51575--51598},
  year={2024}
}

@article{chen2023shikra,
  title={Shikra: Unleashing multimodal llm's referential dialogue magic},
  author={Chen, Keqin and Zhang, Zhao and Zeng, Weili and Zhang, Richong and Zhu, Feng and Zhao, Rui},
  journal={arXiv preprint arXiv:2306.15195},
  year={2023}
}

@inproceedings{you2024ferret,
  title={Ferret: Refer and ground anything anywhere at any granularity},
  author={You, Haoxuan and Zhang, Haotian and Gan, Zhe and Du, Xianzhi and Zhang, Bowen and Wang, Zirui and Cao, Liangliang and Chang, Shih-Fu and Yang, Yinfei},
  booktitle={International Conference on Learning Representations},
  volume={2024},
  pages={57153--57180},
  year={2024}
}

@inproceedings{xia2025bootstrapping,
  title={Bootstrapping grounded chain-of-thought in multimodal llms for data-efficient model adaptation},
  author={Xia, Jiaer and Tong, Bingkui and Zang, Yuhang and Shao, Rui and Zhou, Kaiyang},
  booktitle={Proceedings of the IEEE/CVF International Conference on Computer Vision},
  pages={208--217},
  year={2025}
}

@misc{qwen3.5,
    title  = {{Qwen3.5}: Towards Native Multimodal Agents},
    author = {{Qwen Team}},
    month  = {February},
    year   = {2026},
    url    = {https://qwen.ai/blog?id=qwen3.5}
}

@article{zhang2025pearl,
  title={Perceptual-Evidence Anchored Reinforced Learning for Multimodal Reasoning},
  author={Zhang, Chi and Qiu, Haibo and Zhang, Qiming and Xu, Yufei and Zeng, Zhixiong and Yang, Siqi and Shi, Peng and Ma, Lin and Zhang, Jing},
  journal={arXiv preprint arXiv:2511.18437},
  year={2025}
}

@article{sinha2025chartrvr,
  title={Chart-RVR: Reinforcement Learning with Verifiable Rewards for Explainable Chart Reasoning},
  author={Sinha, Sanchit and Frunza, Oana and Rasul, Kashif and Nevmyvaka, Yuriy and Zhang, Aidong},
  journal={arXiv preprint arXiv:2510.10973},
  year={2025}
}

@article{zhang2026chartr1,
  title={Chart-rl: Generalized chart comprehension via reinforcement learning with verifiable rewards},
  author={Zhang, Xin and Li, Xingyu and Wang, Rongguang and Miao, Ruizhong and Wang, Zheng and Roth, Dan and Li, Chenyang},
  journal={arXiv preprint arXiv:2603.06958},
  year={2026}
}

@inproceedings{selvaraju2019hint,
  title={Taking a hint: Leveraging explanations to make vision and language models more grounded},
  author={Selvaraju, Ramprasaath R and Lee, Stefan and Shen, Yilin and Jin, Hongxia and Ghosh, Shalini and Heck, Larry and Batra, Dhruv and Parikh, Devi},
  booktitle={Proceedings of the IEEE/CVF international conference on computer vision},
  pages={2591--2600},
  year={2019}
}

@inproceedings{hudson2019gqa,
  title={Gqa: A new dataset for real-world visual reasoning and compositional question answering},
  author={Hudson, Drew A and Manning, Christopher D},
  booktitle={Proceedings of the IEEE/CVF conference on computer vision and pattern recognition},
  pages={6700--6709},
  year={2019}
}

@article{das2017vqa_hat,
  title={Human attention in visual question answering: Do humans and deep networks look at the same regions?},
  author={Das, Abhishek and Agrawal, Harsh and Zitnick, Larry and Parikh, Devi and Batra, Dhruv},
  journal={Computer Vision and Image Understanding},
  volume={163},
  pages={90--100},
  year={2017},
  publisher={Elsevier}
}

@article{yuan2025visual,
  title={Visual reasoning tracer: Object-level grounded reasoning benchmark},
  author={Yuan, Haobo and Sun, Yueyi and Li, Yanwei and Zhang, Tao and Deng, Xueqing and Ding, Henghui and Qi, Lu and Wang, Anran and Li, Xiangtai and Yang, Ming-Hsuan},
  journal={arXiv preprint arXiv:2512.05091},
  year={2025}
}

@article{yang2025machine,
  title={Machine mental imagery: Empower multimodal reasoning with latent visual tokens},
  author={Yang, Zeyuan and Yu, Xueyang and Chen, Delin and Shen, Maohao and Gan, Chuang},
  journal={arXiv preprint arXiv:2506.17218},
  year={2025}
}

@article{zhu2026analyzing,
  title={Analyzing Reasoning Consistency in Large Multimodal Models under Cross-Modal Conflicts},
  author={Zhu, Zhihao and Liang, Jiafeng and Jiang, Shixin and Fu, Jinlan and Liu, Ming and Sun, Guanglu and Ng, See-Kiong and Qin, Bing},
  journal={arXiv preprint arXiv:2601.04073},
  year={2026}
}

\appendix

\section{Motivating Diagnostic Details}
\label{app:motivation-diagnostic}

The diagnostic in Figure~\ref{fig:kl-motivation} tests whether hallucinations after RL with only outcome rewards are associated with evidence attention mismatch. We use a Qwen3-VL-4B policy trained with GRPO and final-answer rewards on 903 HallusionBench-Image examples that remain after applying the evidence annotation and filtering pipeline in Appendix~\ref{app:pipeline}. For each generated response, we compute the KL divergence between the response-to-vision attention distribution normalized over visual tokens and the Gaussian mixture evidence target defined in Section~\ref{sec:target}. Responses judged incorrect by the official HallusionBench protocol are treated as hallucinated for this analysis. Figure~\ref{fig:kl-motivation}(a) groups examples into equal-sized KL quantile bins and reports the hallucination rate in each bin, showing whether risk increases as mismatch grows. Spearman correlation is computed over individual examples. Figure~\ref{fig:kl-motivation}(b) compares the KL distributions of hallucinated responses and responses that do not hallucinate, showing whether hallucinated samples have higher mismatch overall. Evidence boxes are used only for this offline diagnostic and are not provided to the model input.

\section{Data Construction Details and Statistics}
\label{app:data}
\label{app:pipeline}

This appendix documents the evidence extraction prompts, box-metadata construction procedure, filtering rules, and dataset-specific preprocessing. Figure~\ref{fig:pipeline} shows the high-level pipeline, and Table~\ref{tab:data-stats} reports the source-level statistics of the annotated pools before final training sampling. The important implementation convention is that evidence boxes are never rendered into the image. Each annotated example stores the original image, question, answer, box coordinates, and a single-/multi-evidence tag. EASE consumes the boxes only when constructing the target attention distribution.

\subsection{Step 1. Evidence Phrase Extraction}

For each source example $(I,q,a)$, GPT-4.1-mini is prompted to enumerate the minimal set of visible entities that must be inspected to verify the answer. The output provides temporary text queries for localization rather than final supervision labels.

\begin{promptbox}{Evidence phrase extraction prompt}
Given the question \(q\) and reference answer \(a\), identify the smallest set of visible and localizable objects or regions that a human would inspect to verify the answer. Include only evidence that directly supports the answer. Return a JSON list of concise noun phrases. Do not include explanations, full sentences, the whole image, generic background, scene-level descriptions, or non-localizable concepts. Return \texttt{[]} if no localizable visual evidence is required.
\end{promptbox}

The extractor returns a structured list of concise noun phrases. It is asked for the smallest set of objects needed to verify the answer, rather than all objects mentioned in the question. We remove phrases referring to the whole scene, the background, the image itself, or non-localizable abstractions, and merge near duplicates by string overlap and head-noun matching.

\subsection{Step 2. Box Localization}

Each intermediate evidence phrase is localized independently, and the resulting coordinates are kept as supervision metadata. We use Grounding DINO~\citep{liu2024groundingdino} as the primary phrase-grounding model and keep its top prediction when the confidence exceeds $\delta_{\mathrm{det}}=0.35$. Candidates below this confidence threshold are discarded.

As a complementary localization signal, we also query a locally deployed Qwen3.5-27B~\citep{qwen3.5} with the original image and the same intermediate phrase to obtain normalized box coordinates. The two localization outputs are merged query by query. If their IoU is at least $0.7$, their union is used as a conservative evidence box $B_k$. If they disagree, the Grounding DINO box is used as the primary localization, and the Qwen box is retained only if it passes Step~3 validation for the same evidence entity. Queries for which neither model returns a valid box are dropped.

\subsection{Step 3. Quality Validation}

The annotation pipeline validates each proposed box independently. Gemini 2.5 Flash-Lite~\citep{comanici2025gemini} receives the original image, question, reference answer, intermediate evidence phrase, and proposed box coordinates, and judges whether the boxed region contains answer-relevant visual content. We remove boxes rejected by the verifier. Using a verifier from a different model family reduces the chance that a single model's extraction bias determines both the intermediate phrase queries and the final boxes.

\paragraph{Independent human audit.}
Three master's students who were not involved in this work independently audit 1,000 sampled instances containing 1,586 evidence boxes. They assess box validity, namely whether a proposed box contains answer-relevant visual evidence, required-evidence coverage, namely whether the proposed set omits evidence needed to verify the answer, and sample-level evidence sufficiency, namely whether the complete box set suffices to verify the answer. The audit yields 95.3\% box precision, 98.6\% required-evidence recall, and 97.8\% sample-level evidence sufficiency, indicating that the automatically constructed evidence boxes provide reliable training-only supervision.

\paragraph{Annotation cost.}
The three-stage pipeline produces approximately 127k evidence-annotated samples. Evidence phrase extraction with GPT-4.1-mini costs approximately \$18, box localization with Grounding DINO and a locally deployed Qwen3.5-27B incurs no external API cost, and box validation with Gemini 2.5 Flash-Lite costs approximately \$30. The total external API cost is therefore approximately \$48.

\begin{promptbox}{Quality validation prompt}
Given the original image, question \(q\), reference answer \(a\), evidence query \(e\), and proposed box coordinates \(b=[x_1,y_1,x_2,y_2]\), decide whether the region specified by \(b\) contains visible evidence matching \(e\) and directly helps verify \(a\). Use the full image for spatial context, but base the final decision on the content inside the proposed box. Return VALID if the box encloses answer-relevant visual evidence. Return INVALID if the box is empty, localizes the wrong object or region, is too vague, or does not help verify the answer. Return only VALID or INVALID.
\end{promptbox}

Boxes are clipped to image boundaries, converted to the coordinate convention used by the visual-token grid, and stored as $\mathcal{B}$. The final output is a metadata-augmented training tuple $(I,q,a,\mathcal{B})$ in which $I$ remains the original unmarked image. During RL rollouts, the policy sees only $(I,q)$, and during the actor update, EASE uses $\mathcal{B}$ to construct the Gaussian-mixture attention target.

\subsection{Training Pool Construction}

After validation, examples with one validated evidence box enter the single-evidence pool, and examples with at least two validated evidence boxes enter the multi-evidence pool. The single-evidence pool provides local grounding supervision, while the multi-evidence pool provides cross-region supervision. Table~\ref{tab:data-stats} summarizes the available annotated examples in each pool. The main EASE training distribution is then sampled from these pools in a balanced 1:1 ratio.

\begin{table*}[t]
\centering
\caption{\textbf{Evidence-annotated data pools.} The table reports source-level annotated examples available after filtering. Counts are measured before final training sampling.}
\label{tab:data-stats}
\begin{tabular}{@{}l@{\hspace{1.0em}}r@{\hspace{1.0em}}r@{\hspace{1.0em}}r@{\hspace{1.3em}}l@{}}
\toprule
\tablehead{Source corpus} & \tableheadtwo{Multi-}{evidence} & \tableheadtwo{Single-}{evidence} & \tableheadtwo{Total}{available} & \tablehead{Primary supervision role} \\
\midrule
ZwZ74K & 0 & 74,000 & 74,000 & Fine-grained local region grounding \\
CLEVR & 4,528 & 1,531 & 6,059 & Controlled compositional reasoning \\
SuperCLEVR & 3,992 & 926 & 4,918 & Harder synthetic multi-object reasoning \\
SpaCE10 & 1,759 & 2,371 & 4,130 & Spatial relations and cross-region evidence \\
ViRL39K & 9,978 & 28,886 & 38,864 & General visual reasoning with localized evidence \\
\midrule
\textbf{Total} & \textbf{20,257} & \textbf{107,714} & \textbf{127,971} & \textbf{Evidence-annotated pool} \\
\bottomrule
\end{tabular}
\end{table*}

Table~\ref{tab:data-stats} reports 20,257 multi-evidence examples from CLEVR~\citep{data_clevr}, SuperCLEVR~\citep{data_super_clevr}, SpaCE10~\citep{data_space10}, and ViRL39K~\citep{vl-rethinker}. It also includes 107,714 single-evidence examples from ZwZ74K~\citep{wei2026zwz} and the same source corpora. The full annotated pool contains 127,971 examples before final training sampling.

\section{Experimental Settings}
\label{app:experimental-settings}

\subsection{Training Datasets}

All controlled runs draw training data from the evidence-annotated VQA pools described in Appendix~\ref{app:data}. The pools are built from ZwZ74K, CLEVR, SuperCLEVR, SpaCE10, and ViRL39K. Each training record keeps the original image, question, reference answer, validated box coordinates, and a tag indicating whether one or multiple evidence boxes are available. The intermediate evidence phrases used for localization are not stored as training fields. Unless otherwise noted, training uses a balanced mixture with equal numbers sampled from the single-evidence and multi-evidence pools. All controlled RL variants use the same sampled image-question-answer data and final-answer verifier. EASE additionally uses the stored box metadata during actor updates to construct its auxiliary attention target.

\subsection{Baselines}

\begin{compactitemize}
\item \textbf{ThinkLite-VL} improves data-efficient visual reasoning through MCTS-guided sample selection~\citep{wang2025thinklite_vl}.
\item \textbf{VL-Rethinker} trains VLMs to reconsider their answers through RL-driven self-reflection~\citep{vl-rethinker}.
\item \textbf{MM-Eureka} studies rule-based RL for multimodal reasoning and releases strong 7B-scale models~\citep{meng2025mm_eureka}.
\item \textbf{NoisyRollout} reinforces visual reasoning with augmented rollouts that perturb the visual input~\citep{liu2025noisyrollout}.
\item \textbf{PAPO$_D$} introduces perception-aware policy optimization to improve reasoning under visual uncertainty~\citep{wang2025papo}.
\item \textbf{VPPO-RL} estimates token-level visual dependence from perturbed images and uses it to modulate policy optimization~\citep{huang2025vppo}.
\item \textbf{VGPO} measures visual focus from internal model states and uses visually guided advantages~\citep{wang2026vgpo}.
\end{compactitemize}

\subsection{Evaluation Benchmarks}

\begin{compactitemize}
\item \textbf{HR-Bench} evaluates high-resolution image perception and fine-grained visual recognition~\citep{hrbench}.
\item \textbf{V*} focuses on guided visual search and localized object understanding~\citep{vstar}.
\item \textbf{CV-Bench} evaluates vision-centric perception skills in two-dimensional and three-dimensional visual contexts~\citep{cvbench}.
\item \textbf{POPE} probes object hallucination through binary visual questions with controlled object presence~\citep{li2023pope}.
\item \textbf{HallusionBench-Image} diagnoses hallucination and visual illusion in image-grounded question answering~\citep{guan2023hallusionbench}.
\item \textbf{MathVista} evaluates mathematical reasoning in visual contexts across diverse problem types~\citep{lu2024mathvista}.
\item \textbf{MathVerse$_{V}$} denotes the vision-centric subset of MathVerse, where solving the problem requires information from the image~\citep{zhang2024mathverse}.
\item \textbf{WeMath} provides a diagnostic evaluation of multimodal mathematical reasoning~\citep{qiao2024wemath}.
\item \textbf{MMK12} covers K--12 multimodal reasoning problems and is used to assess general visual problem solving~\citep{meng2025mm_eureka}.
\item \textbf{LogicVista} evaluates logical reasoning in visual contexts, complementing the perception and mathematical reasoning suites with diagram-grounded reasoning tasks~\citep{xiao2024LogicVista}.
\end{compactitemize}

\subsection{Implementation Details}
\label{app:implementation}

\textbf{Training and evaluation setup.}
Unless otherwise noted, models are trained for 2 epochs with learning rate \(1\times10^{-6}\), rollout batch size 512, and maximum response length 1,024. Evaluation uses greedy decoding with temperature 0.0 for all experiments.

\textbf{EASE configuration.}
Unless otherwise noted, EASE uses the middle-layer attention index \(\ell^\star=\lfloor 2L/3\rfloor\), samples at most 64 valid response tokens for the auxiliary loss, and applies model-to-target KL only to reward-positive trajectories. We set \(\lambda_{\mathrm{attn}}=0.001\) and background smoothing \(\alpha=0.1\).

Algorithm~\ref{alg:ease} summarizes the training procedure used by EASE.

\begin{algorithm}[t]
\footnotesize
\SetAlgoLined
\KwIn{Evidence data $\mathcal{D}$, policy $\pi_\theta$, verifier $R$, layer $\ell^\star$, reward threshold $\tau$, attention weight $\lambda_{\mathrm{attn}}$}
\KwOut{Fine-tuned policy $\pi_\theta$}
\For{\textnormal{each training step}}{
  Sample evidence tuples $(I,q,a,\mathcal{B})$ from $\mathcal{D}$\;
  Roll out grouped responses $y^{(g)}\sim\pi_\theta(\cdot\mid I,q)$ and compute rewards $r^{(g)}=R(y^{(g)},a)$\;
  Map $\mathcal{B}$ to the visual-token grid and build the smoothed Gaussian-mixture target $P_{\mathrm{target}}$ by Eq.~\ref{eq:target}\;
  Compute the base RL actor loss from the grouped rewards\;
  \ForEach{\textnormal{trajectory with } $r^{(g)}\ge\tau$}{
    Sample valid response tokens $\mathcal{S}^{(g)}$ and extract layer-$\ell^\star$ response-to-vision attention\;
    Average attention heads to obtain $P_{\mathrm{model}}^{g,t}$ and compute the KL loss in Eq.~\ref{eq:attn-kl}\;
  }
  Add the reward-gated attention loss in Eq.~\ref{eq:total} to the actor loss\;
  Update $\pi_\theta$\;
}
\caption{\textbf{EASE training procedure.}}
\label{alg:ease}
\end{algorithm}

\section{Reward-Gate Coverage and Prompt Difficulty}
\label{app:gate-coverage}

We estimate initial solvability using eight rollouts from the frozen Qwen3-VL-4B policy and group prompts by their number of successful rollouts. During EASE training, 47.4\% of sampled rollouts are reward-positive and 92.3\% of training groups contain at least one gate-passing rollout. Among prompts initially gated off, 48.6\% later produce at least one positive-reward rollout. Table~\ref{tab:gate-coverage} shows that, although group-level gate coverage is lower for harder prompts, EASE improves over DAPO in every difficulty group.

\begin{table}[t]
\centering
\caption{\textbf{Reward-gate coverage and EASE gains by initial prompt difficulty.} Initial solvability is measured with eight frozen-policy rollouts.}
\label{tab:gate-coverage}
\small
\setlength{\tabcolsep}{2.5pt}
\begin{tabular}{@{}lrrr@{}}
\toprule
\tableheadtwo{Initial}{solvability} & \tableheadtwo{Prompt}{count} & \tableheadtwo{Group gate}{coverage} & \tableheadtwo{$\Delta$(EASE}{-- DAPO)} \\
\midrule
0/8 (Hard) & 1,200 & 78.5\% & 3.4 \\
1--3/8 & 2,600 & 88.7\% & 3.8 \\
4--6/8 & 3,400 & 95.1\% & 2.7 \\
7--8/8 (Easy) & 2,800 & 98.6\% & 1.6 \\
\bottomrule
\end{tabular}
\end{table}

\section{Evidence-Attention Alignment Metrics}
\label{app:attention-metrics}

This appendix defines the diagnostics reported in Figure~\ref{fig:attention-faithfulness}. They measure whether response-to-vision attention is spatially aligned with answer-relevant evidence regions. Evidence boxes are used only for offline analysis and are never provided to the model input.

\paragraph{Response-to-vision attention.}
For each generated response, we extract response-to-vision attention from the same layer used by EASE, \(\ell^\star=\lfloor 2L/3\rfloor\), where \(L\) denotes the number of transformer layers in the language model. For a response token \(t\), attention head \(h\), and visual token \(v\in\mathcal{V}\), we compute
\begin{equation}
p_h^t(v)=
\frac{
\exp((q_{h,t}^{\ell^\star})^\top k_{h,v}^{\ell^\star}/\sqrt{d})
}{
\sum_{v'\in\mathcal{V}}
\exp((q_{h,t}^{\ell^\star})^\top k_{h,v'}^{\ell^\star}/\sqrt{d})
}.
\end{equation}
We average over heads and valid response tokens to obtain a single visual-token distribution:
\begin{equation}
P_{\mathrm{model}}(v)
=
\frac{1}{|\mathcal{S}|}
\sum_{t\in\mathcal{S}}
\frac{1}{H}
\sum_{h=1}^{H}
p_h^t(v),
\end{equation}
where \(\mathcal{S}\) is the set of valid response tokens used for evaluation. \(P_{\mathrm{model}}\) is normalized over the visual-token set \(\mathcal{V}\).

\paragraph{Mapping evidence boxes to visual tokens.}
Let the input image have width \(W_{\mathrm{img}}\) and height \(H_{\mathrm{img}}\), and let the visual-token grid have width \(W_v\) and height \(H_v\). Each visual token \(v_{i,j}\) is assigned an image-plane center
\begin{equation}
c_{i,j}
=
\left(
\frac{j+0.5}{W_v}W_{\mathrm{img}},\,
\frac{i+0.5}{H_v}H_{\mathrm{img}}
\right).
\end{equation}
For an evidence box \(B_k=(x_{1,k},y_{1,k},x_{2,k},y_{2,k})\), we define a binary token mask
\begin{equation}
\begin{aligned}
M_k(v_{i,j})
&=
\mathbb{1}
[
x_{1,k}\le c_{i,j}^{x}\le x_{2,k} \\
&\qquad\land\,
y_{1,k}\le c_{i,j}^{y}\le y_{2,k}
].
\end{aligned}
\end{equation}
The union evidence mask is
\begin{equation}
M_{\mathrm{evid}}(v)
=
\max_{k=1}^{K} M_k(v).
\end{equation}

\paragraph{Evidence Attention Mass.}
Evidence Attention Mass measures the total response-to-vision attention assigned to any annotated evidence region:
\begin{equation}
\mathrm{EAM}
=
\sum_{v\in\mathcal{V}}
P_{\mathrm{model}}(v) M_{\mathrm{evid}}(v).
\end{equation}
Higher values indicate that a larger fraction of visual attention falls inside answer-relevant evidence boxes.

\paragraph{Pointing Accuracy.}
Pointing Accuracy measures whether the strongest attention peak falls inside an evidence region. We first identify the most-attended visual token:
\begin{equation}
v^\star
=
\arg\max_{v\in\mathcal{V}} P_{\mathrm{model}}(v).
\end{equation}
A sample is counted as correct for pointing if \(M_{\mathrm{evid}}(v^\star)=1\). The reported score is the average over evaluation examples:
\begin{equation}
\mathrm{PointingAcc}
=
\frac{1}{N}
\sum_{i=1}^{N}
\mathbb{1}
[
M_{\mathrm{evid}}^{(i)}(v_i^\star)=1
].
\end{equation}

\paragraph{Multi-evidence Coverage.}
Multi-evidence Coverage is computed only on examples with \(K>1\) evidence boxes. For each box \(B_k\), we compute its attention mass
\begin{equation}
m_k
=
\sum_{v\in\mathcal{V}}
P_{\mathrm{model}}(v)M_k(v).
\end{equation}
Because larger boxes naturally cover more visual tokens, we compare \(m_k\) with a size-normalized uniform-attention baseline
\begin{equation}
u_k
=
\frac{\sum_{v\in\mathcal{V}} M_k(v)}
{|\mathcal{V}|}.
\end{equation}
A box is counted as covered if
\begin{equation}
m_k > \beta u_k,
\end{equation}
where we use \(\beta=2\) by default. The coverage score for one multi-evidence example is
\begin{equation}
\mathrm{Coverage}
=
\frac{1}{K}
\sum_{k=1}^{K}
\mathbb{1}[m_k>\beta u_k].
\end{equation}
The reported Multi-evidence Coverage is the average of this quantity over all multi-evidence examples.

\paragraph{Interpretation.}
These metrics operationalize attention-based diagnostics of the evidence-acquisition process targeted by EASE. Evidence Attention Mass measures total evidence alignment, Pointing Accuracy measures whether the strongest attention peak lands on evidence, and Multi-evidence Coverage measures whether attention is distributed across multiple supporting regions rather than concentrated on a single salient region. These diagnostics do not establish causal necessity of the attended regions. Instead, they provide quantitative evidence that EASE increases the spatial alignment between response-to-vision attention and annotated visual evidence.

\section{Quantitative Reasoning Quality Evaluation}
\label{app:reasoning-quality}

We conduct a blind pairwise comparison of EASE and DAPO on 1,000 held-out evidence-annotated instances, retaining only examples for which both final answers are correct under the task verifier. GPT-5.4-mini receives the original image, question, reference answer, annotated evidence, and anonymized responses in randomized order, and assesses evidence-groundedness, visual-claim correctness, hallucination avoidance, and reasoning validity. EASE wins substantially more comparisons than DAPO, providing an output-level assessment of evidence-grounded reasoning quality.

\begin{table}[t]
\centering
\caption{\textbf{Blind pairwise evaluation of reasoning quality.} GPT-5.4-mini compares anonymized EASE and DAPO responses on the answer-correct subset.}
\label{tab:reasoning-quality}
\small
\begin{tabular}{@{}lccc@{}}
\toprule
\textbf{Comparison} & \textbf{EASE win} & \textbf{DAPO win} & \textbf{Tie} \\
\midrule
EASE vs DAPO & 83.3\% & 5.2\% & 11.5\% \\
\bottomrule
\end{tabular}
\end{table}

\section{Responsible Research Details}
\label{app:responsible_details}

\paragraph{Artifact use, licenses, and intended use.}
This work uses existing research artifacts, including VLM checkpoints, VQA training corpora, evaluation benchmarks, grounding models, LLM backends, PyTorch, CUDA, and multimodal RL training code, for research evaluation of evidence-anchored supervision in multimodal RLVR. We cite the original creators of the models, datasets, benchmarks, and baseline systems in the main paper and appendix. We use these artifacts according to the licenses, access conditions, and intended-use requirements specified by their providers. The artifacts produced by our work, including code, prompts, configurations, box metadata, and evaluation scripts, are intended for research on grounded multimodal reasoning, RLVR, and reproducibility analysis, rather than direct safety-critical deployment.


\paragraph{Model-assisted annotation and validation.}
LLMs and VLMs are used as research tools in the evidence annotation pipeline. As described in Appendix~\ref{app:pipeline}, GPT-4.1-mini extracts intermediate evidence phrases, a locally deployed Qwen3.5-27B provides complementary box localization, and Gemini 2.5 Flash-Lite validates proposed box coordinates against the original image, question, answer, and phrase. These model outputs are used only to construct and filter training metadata.

\paragraph{Result reporting.}
We report benchmark accuracy or score, average score across benchmark groups, ablation results, and attention-alignment diagnostics. For controlled comparisons, Base, GRPO, DAPO, and EASE use the same training data, backbone family, verifier, decoding protocol, and evaluation suite within each backbone. 

\paragraph{Use of AI assistants.}
The authors used AI assistants for language polishing and submission-compliance drafting support. All method design, code implementation, experimental design, experimental execution, result analysis, scientific claims, and final writing decisions were conducted, verified, and approved by the authors.

\begin{figure*}[t]
\centering
\includegraphics[width=\textwidth]{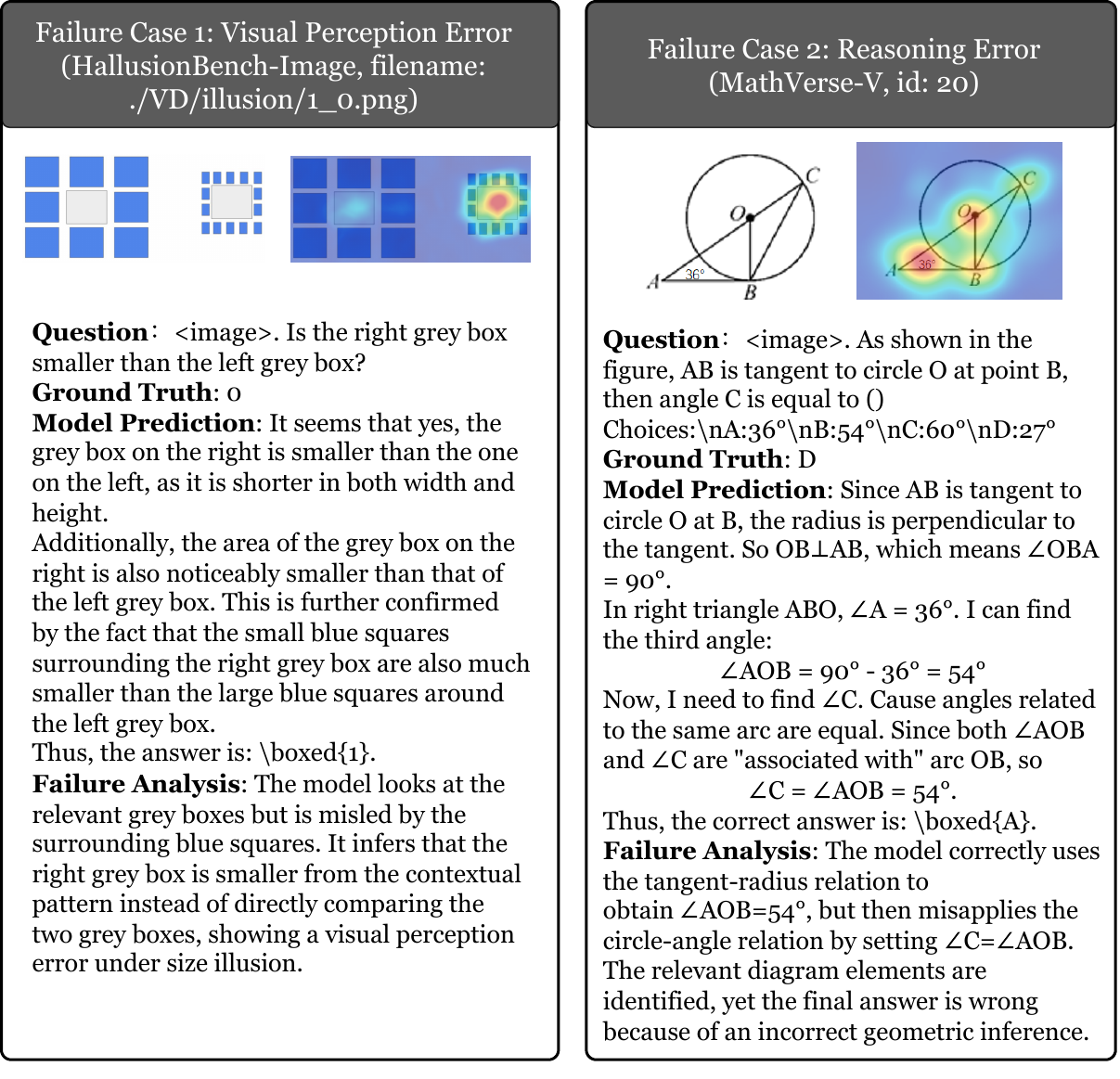}
\caption{\textbf{Representative failure cases.} The left case shows a contextual size illusion from HallusionBench-Image, where the model is misled by surrounding blue squares when comparing the two grey boxes. The right case shows a MathVerse-V geometry example, where the model identifies relevant diagram relations but applies an incorrect circle-angle inference.}
\label{fig:more-qualitative}
\end{figure*}

\section{Failure Case Analysis}
\label{app:qualitative}

To provide deeper insights into the boundaries of our proposed EASE, we present two representative failure cases in Figure~\ref{fig:more-qualitative}. EASE improves evidence acquisition by encouraging response-to-vision attention to align with annotated supporting regions, but correct attention to visual evidence does not guarantee correct recognition or reasoning. We therefore focus on cases where the model attends to the relevant evidence yet still produces an incorrect answer.

\paragraph{Failure Case 1. Visual perception error.}
The first example is a contextual size-illusion case from HallusionBench-Image. The question asks whether the right grey box is smaller than the left grey box, while the correct answer is no. The model attends to the relevant grey boxes but is misled by the surrounding blue squares. It treats the smaller contextual squares around the right box as evidence that the right grey box is also smaller, rather than directly comparing the two grey boxes. This failure suggests that EASE improves where the model looks, but it cannot fully correct a visual representation that is distorted by misleading context.

\paragraph{Failure Case 2. Reasoning after evidence acquisition.}
The second example is a geometry problem from MathVerse-V. The model correctly identifies the tangent-radius relation and derives \(\angle AOB=54^\circ\). However, it then misapplies a circle-angle relation by setting \(\angle C=\angle AOB\), which leads to the incorrect choice. The relevant diagram elements are identified, but the final answer is wrong because the generated solution applies an invalid geometric inference. This failure shows that evidence acquisition is a necessary component of grounded multimodal reasoning, but it is not sufficient when the downstream reasoning policy remains flawed.

Together, these cases complement the positive examples in Figure~\ref{fig:attention-cases}. They indicate that EASE primarily addresses spatial evidence alignment within response-to-vision attention during standard RL, while residual errors can still arise from contextual perception failures or from reasoning mistakes after the relevant evidence has been attended.

\end{document}